\documentclass[final,5p,times,twocolumn,12pt]{elsarticle}

\usepackage{amssymb}
 \usepackage{multirow}
 \usepackage{colortbl}
\journal{}

\begin{document}

\begin{frontmatter}



\title{SS-CPGAN: Self-Supervised Cut-and-Pasting Generative Adversarial Network for Object Segmentation}


\author{Kunal Chaturvedi}

\affiliation{organization={School of Computer Science},%Department and Organization
            addressline={University of Technology Sydney}, 
            city={Ultimo},
            postcode={2007}, 
            state={NSW},
            country={Australia}}

\author{Ali Braytee}
\author{Jun Li}
\author{Mukesh Prasad}
\ead{Kunal.Chaturvedi, Ali.Braytee, Jun.Li, Mukesh.Prasad@uts.edu.au}

\begin{abstract}
This paper proposes a novel self-supervised based Cut-and-Paste GAN to perform foreground object segmentation and generate realistic composite images without manual annotations. We accomplish this goal by a simple yet effective self-supervised approach coupled with the U-Net based discriminator. The proposed method extends the ability of the standard discriminators to learn not only the global data representations via classification (real/fake) but also learn semantic and structural information through pseudo labels created using the self-supervised task. The proposed method empowers the generator to create meaningful masks by forcing it to learn informative per-pixel as well as global image feedback from the discriminator. Our experiments demonstrate that our proposed method significantly outperforms the state-of-the-art methods on the standard benchmark datasets.
\end{abstract}


%end{highlights}


\end{frontmatter}



\section{Introduction}
\label{sec:sample1}

Generative adversarial networks (GANs) \cite{goodfellow2014generative} have become a popular class of image synthesis methods due to their demonstrated ability to create high-dimensional samples with desired data distribution. The primary objective of GANs is to generate diverse, high-quality images while also ensuring the stability of GAN training \cite{brock2018large} \cite{chaturvedi2021automated}. GAN consists of generator and discriminator networks trained in an adversarial manner. The generator attempts to synthesize the real data distribution to fool the discriminator, whereas the discriminator's goal is to distinguish between the generator's real and fake data. In image segmentation, several compositional generative models have been proposed~\cite{chen2019unsupervised,bielski2019emergence,abdal2021labels4free,arandjelovic2019object, ZHANG2022123,ZHAN2022127844}, where the generator creates a synthesized composite image by copying the object from one image and pasting it in another to fool the discriminator about thinking the synthesized composite image is real. But, the generator may not perform any segmentation, and the background may look realistic. Therefore, for effective training, the discriminator must provide the generator with informative learning signals by learning relevant semantics and structures of the data that may result in more effective generators. However, the current state-of-the-art GANs \cite{ZHAN2022127844,rs14051206,LI2022738,8489047,8489550} employ discriminators based on the classification network, which learn only a single discriminative signal such as the difference between real and fake images. In such a non-stationary environment, the generator becomes prone to catastrophic forgetting and may lead to training instability or mode collapse~\cite{chen2019self}.

%talk about methods propsoed as self-supevised GANs

To address the issues above, additional discriminatory signals are required to guide the training mechanism and assist the generator in producing high-quality images. This can be accomplished by increasing the capacity of the discriminator with auxiliary tasks and signals. These auxiliary tasks on the labeled datasets resist the forgetting issues and improve the training stability of GANs, but it suffers with unlabeled datasets. Recently, self-supervised learning has been explored on numerous GANs methods {\cite{chen2019self,patel2021lt,huang2020fx,hou2021self}. The self-supervised tasks provide the learning environment with additional guidance to the standard training mechanism. Most of the recent self-supervision methods on GANs use auxiliary tasks on transformation. For example, SS-GAN developed by Chen et al. \cite{chen2019self} uses rotation prediction as an auxiliary task. In FX-GAN, Huang et al. \cite{huang2020fx} use the pretext task of prediction on corrupted real images, and in LT-GAN \cite{patel2021lt}, the authors use distinguishing GAN-induced transformation as a pretext task. However, the goals of these self-supervised transformation tasks need to be consistent with the GAN's goal of mimicking the real data distribution. Moreover, this problem amplifies when the generator's task is to construct segmentation masks from the foreground images.

Recent self-supervised learning methods have demonstrated remarkable promise in global tasks such as image classification by training simple classifiers on features learned through instance discrimination. However, the pre-text tasks of these global feature-learning approaches do not explicitly retain spatial information, making them unsuitable for object segmentation~\cite{9808406,https://doi.org/10.1111/mice.12674,LIU20213885,10.1145/3554727}. To maintain an enriched real data representation and improve the quality of generated segmentation masks, we propose a \textit{Self-Supervised Cut-and-Paste GAN} (SS-CPGAN) using U-net architecture \cite{ronneberger2015u}, which unifies cut-and-paste adversarial training with a self-supervised task. It allows the discriminator to learn local and global differences between real and fake data. In contrast to the existing transformation self-supervision methods, our self-supervision learning method creates pseudo labels using unsupervised segmentation methods. Then, it simultaneously forces the discriminator to provide the generator with global feedback (real or fake) and the per-pixel feedback of the synthesized images with the help of pseudo labels.

To sum up, the contributions of this paper are as follows:

\begin{itemize}
  \item This paper proposes a novel Self-Supervised Cut-and-Paste GAN (SS-CPGAN), that unifies cut-and-paste adversarial training with a segmentation self-supervised task. SS-CPGAN leverage unlabeled data to maximize segmentation performance and generate highly realistic composite images.
  \item The proposed self-supervised task in SS-CPGAN improves the discriminator’s representation ability by enhancing structure learning with global and local feedback. This enables the generator with additional discriminatory signals to achieve superior results and stabilize the training process.
\item This paper comprehensively analyses the benchmark datasets and compares the proposed method with the baseline methods.

\end{itemize}

%%%%%%%%%%%%%%%%%%%%%%%%%%%%%%%%%%%%%%%%%%
\section{Related Works}
\subsection{Unsupervised Object Segmentation via GANs}
Unsupervised segmentation using GANs is an important topic in research. Several works~\cite{chen2019unsupervised,bielski2019emergence,abdal2021labels4free,arandjelovic2019object} investigate the use of compositional generative models to obtain high-quality segmentation masks. Copy-pasting GAN~\cite{arandjelovic2019object} performs unsupervised object discovery by extracting foreground objects and then copying and pasting them onto different backgrounds. Similarly, PerturbGAN~\cite{bielski2019emergence} generates a foreground mask along with a background and foreground image in an adversarial manner. Recently, Abdal et al. (2021)~\cite{abdal2021labels4free} propose using an alpha network that includes two pre-trained generators and a discriminator on the StyleGAN to generate high-quality masks. These methods learn object segmentation without needing to use annotations. However, they are prone to degenerate solutions or other trivial cases. For example, the generator may not perform any segmentation, and the background looks realistic, or the generator may segment foreground masks consisting of all-ones. To avoid such problems, special care must be taken while training the compositional generative models. Copy-pasting GAN uses anti-shortcut, border-zeroing, blur, and grounded fakes to prevent trivial solutions~\cite{arandjelovic2019object}. PerturbGAN avoids such solutions by randomly shifting object segments relative to the background~\cite{bielski2019emergence}. However, Abdal et al. (2021)~\cite{abdal2021labels4free} make several changes to the original StyleGAN and use a truncation trick along with regularization to avoid degenerate solutions. While these methods achieve object segmentation of foreground objects~\cite{9616450}, the generated segmentation masks are often inferior in quality. Furthermore, due to such non-trivial procedures, training of GANs becomes very challenging and complex \cite{Yang1234}.

\subsection{Self-supervised learning} Self-supervised learning belongs to unsupervised learning, which learns useful feature representations from unlabeled data with the help of pretext tasks. It helps reduce the enormous data collection and annotation cost~\cite{9803869,app12189213}. The traditional way to do this is to give the model some pretext tasks to solve. In this way, the networks learn good feature representations with the help of pseudo labels created by the pretext tasks \cite{9770283}. Recently, many pretext tasks and adversarial training have been introduced~\cite{chen2019self,huang2020fx,patel2021lt,9869893,baykal2020deshufflegan}. The motivation for using self-supervised learning in GANs is to: (1) prevent discriminator forgetting~\cite{thanh2020catastrophic}; (2) improve training stability~\cite{mao2019mode}; (3) and ensure high quality of images generated~\cite{tran2019self}. The self-supervision techniques rely on pretext tasks on geometric transformations (e.g., prediction on rotated images\cite{chen2019self}, corrupted images~\cite{huang2020fx}, GAN-induced transformations~\cite{patel2021lt}, clustering representations~\cite{9869893}, or a deshuffling task that predicts the shuffled orders~\cite{baykal2020deshufflegan}) to increase the discriminator's representation power. These self-supervised tasks may not work well for segmentation due to the inherent differences between the classification and segmentation tasks. In addition to this, image generation for segmentation requires GANs to capture contextual information between the foreground object and the background, which can be complicated in the absence of relevant visual representations. Unlike the aforementioned methods, we incorporate segmentation using self-supervised learning coupled with the Cut-and-Paste GAN to obtain high-quality segmentation masks. Most importantly, with our self-supervised approach, no extra care is needed to deal with the trivial solutions prevalent in compositional generative models. 

%%%%%%%%%%%%%%%%%%%%%%%%%%%%%%%%%%%%%%%%%

\section{Method}
In this section, we first present the standard terminology of adversarial training and the encoder-decoder discriminator. We then introduce our SS-CPGAN method built upon the cut-and-paste adversarial training. The unified framework with the segmentation using self-supervised task encourages the generator to emphasize local and global structures while synthesizing masks.

\subsection{Adversarial Training}
As shown in Figure~\ref{sscpgan}, we build a generative model in which the generator takes the foreground image as the input and generates a composite image using a combination of the predicted mask, source foreground image and the background image to fool the discriminator. Formally, we define the input foreground source image as $I_{f} \in P_{data}$ and the background image as $I_{b} \in P_{data}$ where $P_{data}$ denotes the set of input images. Now, we define a generator ($G$) that is trained in an adversarial manner against the discriminator ($D$). During the training process, the generator predicts a segmentation mask  defined by $m_{g}(I_{f}) \in [0,1]$. Then, using the predicted mask: $m_{g} (I_{f})$, foreground source image: $I_{f}$, and resized background image: $I_{b}$, we define composite image as follows

% $m_{g}(I_{f}) = G(I_{f})$ where $m_{g} (I_{f}) \in [0,1]$. Then, using the predicted mask $m_{g} (I_{f})$, foreground source image $I_{f}$, and background image $I_{b}$, we define composite image as follows

\begin{equation}
I_{C}=m_{g}(I_{f})I_{f} + (1-m_{g}(I_{f}))I_{b}
\label{eq:ic}
\end{equation}

The discriminator's objective is to classify the composite image as real or fake. As a result, the standard objective of the discriminator and the generator of the CPGAN is defined as follows,

\begin{equation}
\mathcal{L}_{D}=E\left[ \log D(I_{f}) + \log(1-D(I_{C})) \right]
\label{eq:LD}
\end{equation}

% \begin{equation}
% \mathcal{L}_{D}=\max_{D} E\left[ \log D(I_{f}) + \log(1-D(I_{C})) \right]
% \label{eq:LD}
% \end{equation}

\begin{equation}
\mathcal{L}_{G}= - E\left[ \log D(I_{C})  \right]
\label{eq:LG}
\end{equation}
The discriminator works as a classification network restricted to learning only through the discriminative differences between the real and fake samples. Thus, the discriminator fails to provide any useful information to the generator. Therefore, we use an encoder-decoder discriminator network with self-supervised learning to mitigate this problem.

\begin{figure*}
\centering
\includegraphics[scale=0.28]{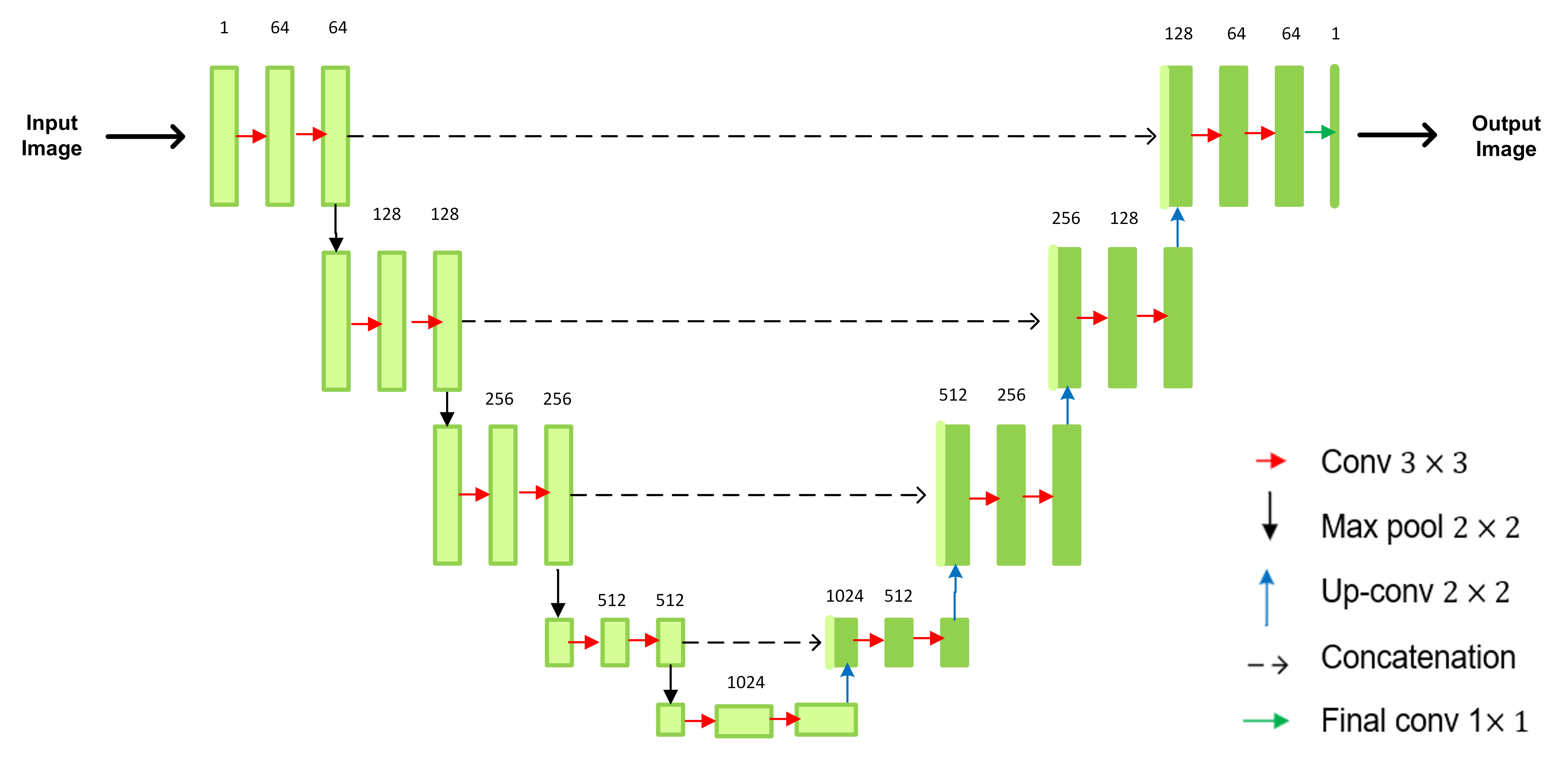}
\caption{An overview of U-net architecture. The different arrows denote the different operations used in the encode-decoder architecture.}
\label{unet}
\end{figure*}

\subsection{Encoder-Decoder Discriminator}
In this work, we replace the standard classification discriminator with the U-net based discriminator. The U-net is an encoder-decoder architecture that consists of a network of convolutional layers, and skip connections for semantic segmentation~\cite{10018569,10.3389/fenvs.2022.996513,9810920}. It was initially proposed for biomedical image segmentation, which achieved precise segmentation results with few training images. Further, it demonstrates good results in other applications, including geo-sciences~\cite{chen2020mapping}, remote sensing~\cite{tran2019robust}, and others. Its architecture (see Figure~\ref{unet}) is symmetric and consists of two paths, an Encoder that extracts spatial features from the input image (downscaling process), and a Decoder that constructs the segmentation map from the extracted feature maps (upscaling process). The use of U-net architecture in the proposed model adds major advantages: 1) it enables simultaneous use of the global location and context while predicting masks; 2)  it retains the full context of the input images, which is a significant advantage over patch segmentation approaches \cite{9053405}.

We use the encoder part of the U-net as the standard classification discriminator that performs the binary decision on real/fake composite images. And the decoder part of the U-net architecture is utilized by the self-supervised task to give per-pixel feedback on the synthesized images with the help of pseudo labels. This allows the discriminator to learn relevant local and global differences between real/fake images.

\begin{figure*}
\centering
\includegraphics[scale=0.7]{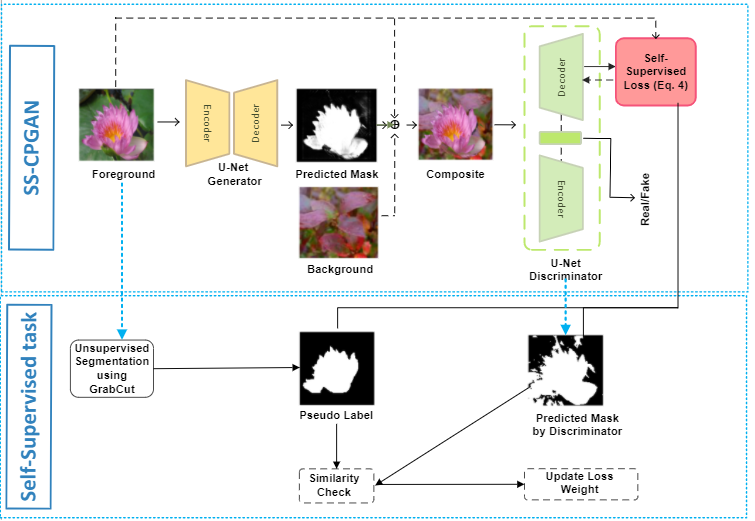}
\caption{The proposed Self-supervised cut-and-paste GAN (SS-CPGAN)}
\label{sscpgan}
\end{figure*}

\subsection{Self-Supervised Cut-and-Paste GAN (SS-CPGAN)}
To improve the representation learning ability of the CPGAN, the discriminator must learn semantic and structural information from the synthesized images. Therefore, we use self-supervised learning to build comprehensive representations for the CPGAN. In this work, we employ a segmentation self-supervised task, to enable the discriminator with enhanced learned features that ultimately empower the generator to create consistent and structurally coherent masks. The pseudo segmentation masks $m_{US} (I_{f} ) \in [0,1]$ are created using a graph unsupervised segmentation algorithm. These masks obtained by the GrabCut technique act as a suitable prior for the U-net discriminator (see, Figure~\ref{sscpgan} (top)). Here, the discriminator performs two important tasks, i.e., (1) classification of real/fake compositing images; and (2) performing per-pixel classification on $I_{f} \in P_{data}$ to generate segmentation masks. Given the self-supervised pseudo labels, we train the discriminator for accurate pixel-level prediction. Integrating self-supervisory signals empowers the discriminator by enhancing its localization ability and forces it to learn useful semantic representations. This mechanism enables the generator to achieve optimized results and makes the training process more stabilized. 

Formally, we define $I_{f} \in P_{data}$ as the source image containing the foreground object, and $P_{data}$ denotes the set of input images. Further, we create a pseudo label denoted by $m_{US} (I_{f} )  \in [0,1]$, using an unsupervised segmentation algorithm. Then, we define $m_{w} (I_{C} )\in [0,1]$ as the pixel-wise segmentation mask produced by the decoder of the discriminator. Hereafter, we optimize the overall discriminator loss function (Eq.~\ref{eq:LDprime}) by augmenting a new self-supervision loss (Eq.~\ref{eq:LSS}) 

\begin{equation}
\mathcal{L}_{self-supervised}=\mathcal{L}\left( m_{w}(I_{C}), 1-m_{US}(I_{f}) \right)
\label{eq:LSS}
\end{equation}

\begin{equation}
\mathcal{L'}_{D}=\mathcal{L}_{D} + \lambda \mathcal{L}_{self-supervised} 
\label{eq:LDprime}
\end{equation}
where $\mathcal{L}$ is the cross-entropy loss, and $\lambda$ denotes the loss weight for the self-supervision loss. This hyperparameter is updated according to the comparison between $m_{US} (I_f)$ and $m_{w} (I_{C})$, using intersection-over-union (IoU). The details of the hyperparameter chosen are explained in the implementation details section. The framework of self-supervised learning is shown in Figure~\ref{sscpgan} (bottom).

%%%%%%%%%%%%%%%%%%%%%%%%%%%%%%%%%%%%%%%%%

\section{Experimentation}
This section discusses the implementation details of the proposed method and an extensive set of experiments on various datasets.

\subsection{Datasets}

We utilize five different datasets for the foreground and background set to train our SS-CPGAN as described below:

\begin{itemize}
\item \textbf{Caltech-UCSD Birds (CUB) 200-2011} is a frequently used benchmark for unsupervised image segmentation. It consists of 11,788 images from 200 bird species.

\item \textbf{Oxford 102 Flowers} consists of 8,189 images from 102 flower classes.

\item \textbf{FGVC Aircraft (Airplanes)} contains 102 different aircraft model variants with 100 images of each. This dataset was used initially for fine-grained visual categorization.

\item \textbf{MIT Places2} is a scene-centric dataset with more than 10 million images consisting of over 400 unique scene classes. However, in the experiments, we use the classes: rainforest, forest, sky, and swamp as a background set for the Caltech-UCSD Birds dataset, and in the Oxford 102 Flowers, we use the class: herb garden as a background set.

\item \textbf{Singapore Whole-sky IMaging CATegories (SWIMCAT)} contains 784 images of five categories: patterned clouds, clear sky, thick dark clouds, veil clouds, and thick white clouds. We use the SWIMCAT dataset as a background set for the FGCV dataset.

\end{itemize}

We chose background datasets similar to the background of the images from the foreground dataset. For the foreground datasets, we use Caltech-UCSD Birds (CUB) 200-2011 \cite{welinder2010caltech}, Oxford 102 Flowers \cite{nilsback2008automated}, and FGCV Aircraft (Airplanes) \cite{maji2013fine}. During the training, we do not utilize the masks available with datasets, Caltech-UCSD Birds and Flowers-102. For the background datasets, we use MIT Places2 \cite{zhou2017places}, and SWIMCAT \cite{dev2015categorization}.
}

\subsection{Experimental Setings}
Our implementation used the PyTorch framework. For training our models, we deploy a batch size of $16$ and the Adam optimizer with an initial learning rate of $2.10^{-4}$. The training images are reshaped to $64\times 64$, $128\times 128$, and $256\times 256$. For the self-supervision task, we use the GrabCut technique \cite{rother2004grabcut} as the unsupervised segmentation algorithm. 

\begin{figure*}[t]
\centering
\includegraphics[scale=0.9]{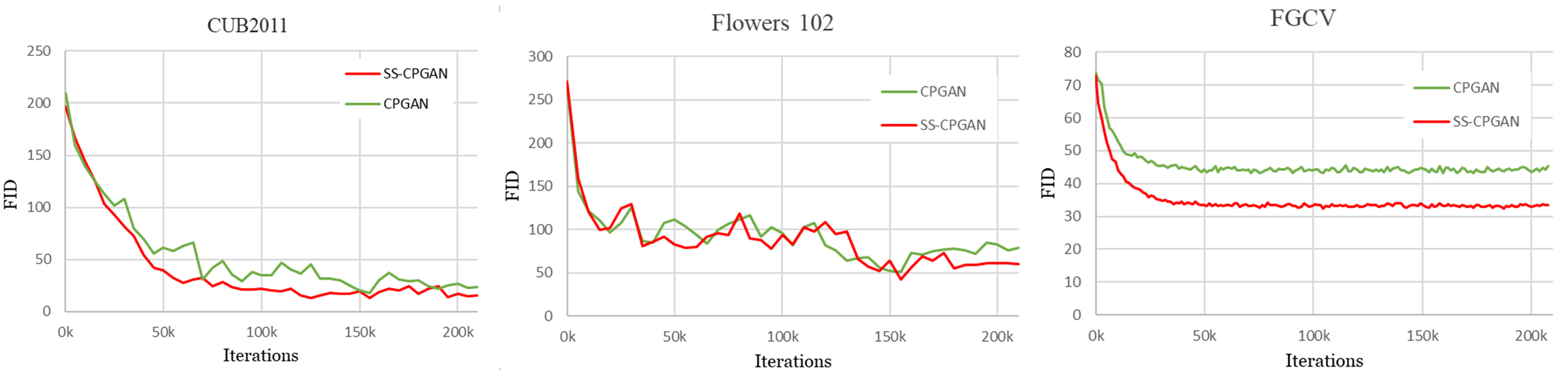}
\caption{FID training curves for CPGAN and SS-CPGAN on the datasets: CUB2011, Flowers 102, and FGCV.}
\label{fig1}
\end{figure*}

\subsection{Hyper-parameter Range}

SS-CPGAN presents a new self-supervised loss, i.e. $\mathcal{L}_{self-supervised}$ that need to be validated. As shown in Table \ref{tbl:SSIM}, we present Structural similarity (SSIM) scores according to different values of $\lambda$. The SSIM scores vary between the range of [0,1], with lower values indicating the lower quality of generated images. During the experiments, we find that the optimal values of the hyper-parameter can vary depending on the Intersection-Over-Union (IoU) score between the pseudo label (mask) and the predicted mask by the discriminator. Initially, when IoU $< 0.2$, the hyperparameter value is set to $0.5$ to boost the model's ability to learn useful representations from the pseudo label. When the $0.2 <$ IoU $< 0.8$, we refine the predicted mask using the hyperparameter value $\lambda$ of $0.1$. To avoid the pseudo labels compromising the predicted masks, we restrict the value $\lambda$ to $0$ when the IoU $> 0.8$.

\begin{table}[htb]
\setlength{\tabcolsep}{3.5pt} % Default value: 6pt
\centering
\caption{Validation of hyper-parameter choices for $\lambda$ in the self-supervised loss}
\label{tbl:SSIM}
\begin{tabular}{l|l|l|l|l|l} 
\hline
\multicolumn{6}{c}{SSIM $\uparrow$}                       \\ \hline
lambda & 0.1   & 0.5   & 1     & 10    & 100   \\ \hline
SSIM   & 0.657 & 0.834 & 0.450 & 0.421 & 0.125 \\ \hline
\end{tabular}
\end{table}

\begin{table*}
\setlength{\tabcolsep}{4pt} % Default value: 6pt
\centering
\caption{FID comparison of the proposed method with the baseline CPGAN model}
\label{tbl:fid}
\begin{tabular}{l|l|l|l|l} 
\hline
\multicolumn{5}{l}{~ ~ ~ ~ ~ ~ ~ ~ ~ ~ ~ ~ ~ ~ ~ ~ ~ ~ ~ ~ ~ ~FID $\downarrow$}                                                                                                                                                            \\ 
\hline
Methods           & Image size & \begin{tabular}[c]{@{}l@{}}Caltech \\UCSD-\\Bird 200\end{tabular} & \begin{tabular}[c]{@{}l@{}}FGCV- \\Aircraft\end{tabular} & \begin{tabular}[c]{@{}l@{}}Oxford\\102\\Flowers\end{tabular}  \\ 
\hline
CPGAN             & 64 x 64    & 26.724                                                            & 43.353                                                   & 81.724                                                        \\
                  & 128 x 128  & 23.002                                                            & 39.674                                                   & 44.825                                                        \\
                  & 256 x 256  & 21.346                                                            & 44.825                                                   & 51.218                                                        \\ 
\hline
\textbf{SS-CPGAN} & 64 x 64    & \textbf{22.342}                                                 & \textbf{39.578}                                          & \textbf{63.343}                                               \\
                  & 128 x 128  & \textbf{15.634}                                                   & \textbf{37.756}                                          & \textbf{54.982}                                               \\
                  & 256 x 256  & \textbf{13.113}                                                   & \textbf{33.149}                                          & \textbf{49.181}                                              
\end{tabular}
\label{tbl:fid}
\end{table*}

\subsection{Results}
We utilized the Fréchet Inception Distance (FID) score and mean Intersection over Union (mIoU) metric for the quantitative evaluation of our methods. In this work, we use the FID score on the datasets CUB2011, Oxford 102 Flowers, and FGCV Aircraft (see Table~\ref{tbl:fid}) to compare the SS-CPGAN model with the CPGAN model images spatially scaled to $64 \times 64$, $128 \times 128$, and $256 \times 256$. For the datasets with available ground truth masks, including CUB2011, and Oxford 102 Flowers, we use the mIoU metric as shown in Table~\ref{tbl:miou}.

\begin{figure*}
\centering
\includegraphics[scale=0.8]{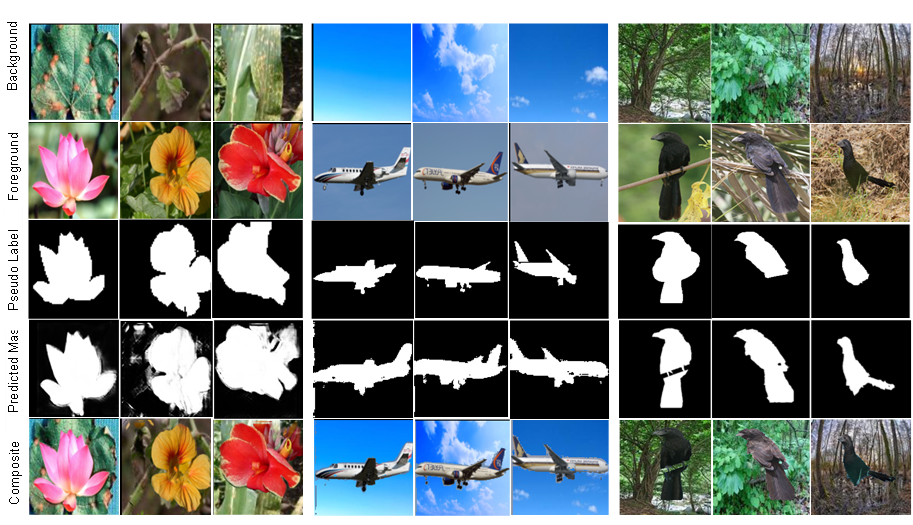}
\caption{Visualization results with the proposed SS-CPGAN on the datasets: Oxford 102 Flowers (left), FGVC Aircraft (center), and Caltech-UCSD Birds (CUB) 200-2011 (right).}
\label{testimages}
\end{figure*}

In Figure~\ref{fig1}, we report the FID scores over the training iterations. We show that our method stabilizes GAN training across all the datasets by allowing GAN training to converge faster and consistently improve performance throughout the training. According to Figure~\ref{fig1}, our method, SS-CPGAN, utilizing self-supervision outperforms the baseline method, CPGAN, on each dataset used. Furthermore, as shown in Figure~\ref{testimages}, the generated masks and composite images of our proposed SS-CPGAN are of superior quality. The standard classification discriminator of CPGAN does not provide effective guidance to the generator. During the training, the standard discriminator is not encouraged to learn a more robust data representation. The classification task learns only the representation based on the discriminative differences between real/fake images and fails to give information on why the synthesized image looks fake. Notedly, our self-supervision task assigned to the U-net discriminator provides the generator with global feedback (real or fake) and per-pixel feedback of the masks with the help of pseudo labels. The self-supervisory signals prevent the two scenarios for the generator which the standard discriminator fails to do, i.e., creating constant masks of only all-zeros pixel values or all-ones pixel values. The enhanced discriminator of SS-CPGAN influences the generator to create high quality masks that are devoid of any such anomalies. As shown in Figure~\ref{testimages}, the qualitative analysis of the proposed SS-CPGAN shows that the generated masks and composite images are of superior quality.

\begin{table}
\setlength{\tabcolsep}{3.5pt} % Default value: 6pt
\centering
\caption{mIOU comparison of the proposed method with the baseline CPGAN model}
\label{tbl:miou}
\begin{tabular}{l|l|l|l} 
\hline
\multicolumn{4}{l}{~ ~ ~ ~ ~ ~ ~ ~ ~ ~ ~ ~ ~ ~ ~ ~ ~ ~ ~ ~ ~ ~mIoU $\uparrow$}                                                                                                        \\ 
\hline
Methods                   & Image size & \begin{tabular}[c]{@{}l@{}}Caltech \\UCSD-\\Bird 200\end{tabular} & \begin{tabular}[c]{@{}l@{}}Oxford\\102\\Flowers\end{tabular}  \\ 
\hline
w/o Self-Supervision      & 64 x 64    & 0.537                                                             & 0.632                                                         \\
                          & 128 x 128  & 0.492                                                             & 0.674                                                         \\
                          & 256 x 256  & 0.484                                                             & 0.779                                                         \\ 
\hline
\textbf{Self-Supervision} & 64 x 64    & \textbf{0.571}                                                    & \textbf{0.625}                                                \\
                          & 128 x 128  & \textbf{0.543}                                                    & \textbf{0.719}                                                \\
                          & 256 x 256  & \textbf{0.518}                                                    & \textbf{0.791}                                               
\end{tabular}
\label{fig1}
\end{table}

\subsection{Comparison with the state-of-the-art}
We compare our self-supervision based Cut-and-Paste GAN (SS-CPGAN) with state-of-the-art. As shown in Table~\ref{tbl:cfid}, we report and compare the FID score on the Caltech UCSD-Bird 200 dataset. Specifically, the FID scores of StackGANv2~\cite{zhang2018stackgan}, OneGAN \cite{benny2020onegan}, LR-GAN \cite{yang2017lr}, ELGAN \cite{yang2021unsupervised}, and FineGAN~\cite{singh2019finegan} are listed. The results in Table~\ref{tbl:cfid} show that our method delivers better performance and outperforms the existing methods. LR-GAN \cite{yang2017lr} performed the worst, followed by the other methods. The low performance of layer-wise GANs \cite{yang2017lr} \cite{yang2021unsupervised} is attributed to the fact that these methods are prone to degenerate during the training phase, with all the pixels being assigned as one component. In Table~\ref{tbl4}, we compare the performance of our method to the recent methods using the mIoU metric on Caltech UCSD-Bird 200 and Oxford flowers-102 respectively. In comparison to PerturbGAN \cite{bielski2019emergence}, ContraCAM \cite{mo2021object}, ReDO \cite{chen2019unsupervised}, UISB \cite{kim2020unsupervised} and IIC-seg \cite{ji2019invariant}, our method outperforms by a large margin on Caltech UCSD-Bird 200 dataset. On the Oxford flowers-102 dataset, we perform better than the methods, ReDO \cite{chen2019unsupervised}, Kyriazi et. al~\cite{melas2021finding} and Voynov et. al.~\cite{voynov2021object}. Here, ReDO and Kyriazi et. al (2021) are unsupervised approaches whereas Voynov et. al (2021) is a weakly supervised approach to create segmentation maps.  The ability to leverage pseudo labels in the training of Cut-and-Paste GAN assists in creating foreground masks of superior quality.

\begin{table}[htb]
\centering
\caption{FID comparison of our proposed method SS-CPGAN with the state-of-art on Caltech UCSD-Bird 200 dataset}
\label{tbl:fid}
\begin{tabular}{l|l} 
\hline
Method            & FID             \\ 
\hline
StackGANv2        & 21.4            \\
FineGAN           & 23.0            \\
OneGAN            & 20.5            \\
LR-GAN            & 34.91           \\
ELGAN             & 15.7            \\
\textbf{SS-CPGAN} & \textbf{13.11} 
\end{tabular}

\label{tbl:cfid}
\end{table}

\begin{table}[htb]
\centering
\caption{Quantitative comparison of the segmentation performance of our method SS-CPGAN with the state-of-art}
\label{tbl4}
\begin{tabular}{l|l|l} 
\hline
Dataset               & Method            & mIoU            \\ 
\hline
Caltech UCSD-Bird 200 & PerturbGAN        & 0.380           \\
                      & ContraCAM         & 0.460           \\
                      & ReDO              & 0.426           \\
                      & UISB              & 0.442           \\
                      & IIC-seg           & 0.365           \\
                      & \textbf{SS-CPGAN} & \textbf{0.571}  \\ 
\hline
Oxford 102 flowers    & ReDO              & 0.764           \\
                      & Kyriazi et. al.   & 0.541           \\
                      & Voynov et al.     & 0.540           \\
                      & \textbf{SS-CPGAN} & \textbf{0.791} 
\end{tabular}
\end{table}

%%%%%%%%%%%%%%%%%%%%%%%%%%%%%%%%%%%%%%%%

\section{Conclusion}
In this work, we proposed a novel Self-Supervised Cut-and-Paste GAN method to learn object segmentation. Specifically, we unify the cut-and-paste adversarial training with the proposed segmentation based self-supervision learning. Unlike the existing transformation self-supervised methods, our method improves the discriminator's representation ability by enhancing structure learning with global and local feedback from the synthesized masks. Furthermore, SS-CPGAN overcomes the issue of unwanted trivial solutions (generating constant masks of only all-zeros or all-ones pixel values) that plagues the generator. The experimental results show that our approach generates superior quality images and achieves promising results on the benchmark datasets.

%\appendix

%\section{Sample Appendix Section}
%Lorem ipsum dolor sit amet, consectetur adipiscing elit, sed do eiusmod tempor section \ref{sec:sample1}  

 \bibliographystyle{elsarticle-num} 
 \bibliography{cas-refs}





% \end{thebibliography}
\end{document}